\def\paperlanguage{}
\pdfoutput=1 

\documentclass[letterpaper, 10 pt, conference]{ieeeconf}  

\usepackage{bm}
\usepackage{cite}
\usepackage{flushend}
\include{preamble}

\IEEEoverridecommandlockouts                              

\title{\LARGE \textbf
  {
    \switchlanguage%
    {%
      Environmentally Adaptive Control Including Variance Minimization Using Stochastic Predictive Network with Parametric Bias:\\Application to Mobile Robots
    }%
    {%
      確率的RNNPBによる分散最小化を含む環境適応型制御: 台車型ロボットへの適用
    }%
  }
}

\author{Kento Kawaharazuka$^{1}$, Koki Shinjo$^{1}$, Yoichiro Kawamura$^{1}$, Kei Okada$^{1}$, and Masayuki Inaba$^{1}$
  \thanks{$^{1}$ The authors are with the Department of Mechano-Informatics, Graduate School of Information Science and Technology, The University of Tokyo, 7-3-1 Hongo, Bunkyo-ku, Tokyo, 113-8656, Japan.
    {\texttt\small [kawaharazuka, shinjo, y-kawamura, k-okada, inaba]@jsk.t.u-tokyo.ac.jp}
  }
}

\begin{document}

\maketitle
\thispagestyle{empty}
\pagestyle{empty}

\begin{abstract}
  \switchlanguage%
  {%
    In this study, we propose a predictive model composed of a recurrent neural network including parametric bias and stochastic elements, and an environmentally adaptive robot control method including variance minimization using the model.
    Robots which have flexible bodies or whose states can only be partially observed are difficult to modelize, and their predictive models often have stochastic behaviors.
    In addition, the physical state of the robot and the surrounding environment change sequentially, and so the predictive model can change online.
    Therefore, in this study, we construct a learning-based stochastic predictive model implemented in a neural network embedded with such information from the experience of the robot, and develop a control method for the robot to avoid unstable motion with large variance while adapting to the current environment.
    This method is verified through a mobile robot in simulation and to the actual robot Fetch.
  }%
  {%
    本研究では, parametric biasの仕組みと確率的要素を含むrecurrent neural networkにより構成された予測モデルの構築と, それを使った分散最小化を含む環境適応型ロボット制御手法について提案する.
    柔軟な身体を持つロボットや部分的にしか状態を観測できないロボットはモデル化が難しく, その予測モデルは確率的な振る舞いをする場合が多い.
    また, ロボットの身体状態や動作する環境は逐次的に変化し, その予測モデルが変化してしまう場合がある.
    本研究では, 実機における動作データから, これらの情報を埋め込んだニューラルネットワークを構築し, 現在の環境に適応しつつ, 分散が大きく不安定な動作を回避してロボットが動くための動作制御を開発する.
    本手法を, 台車のシミュレーション・単腕mobile robot Fetchの足回りに適用し, その有効性を確認する.
  }%
\end{abstract}

\section{Introduction}\label{sec:introduction}
\begin{figure}[t]
  \centering
  \includegraphics[width=1.0\columnwidth]{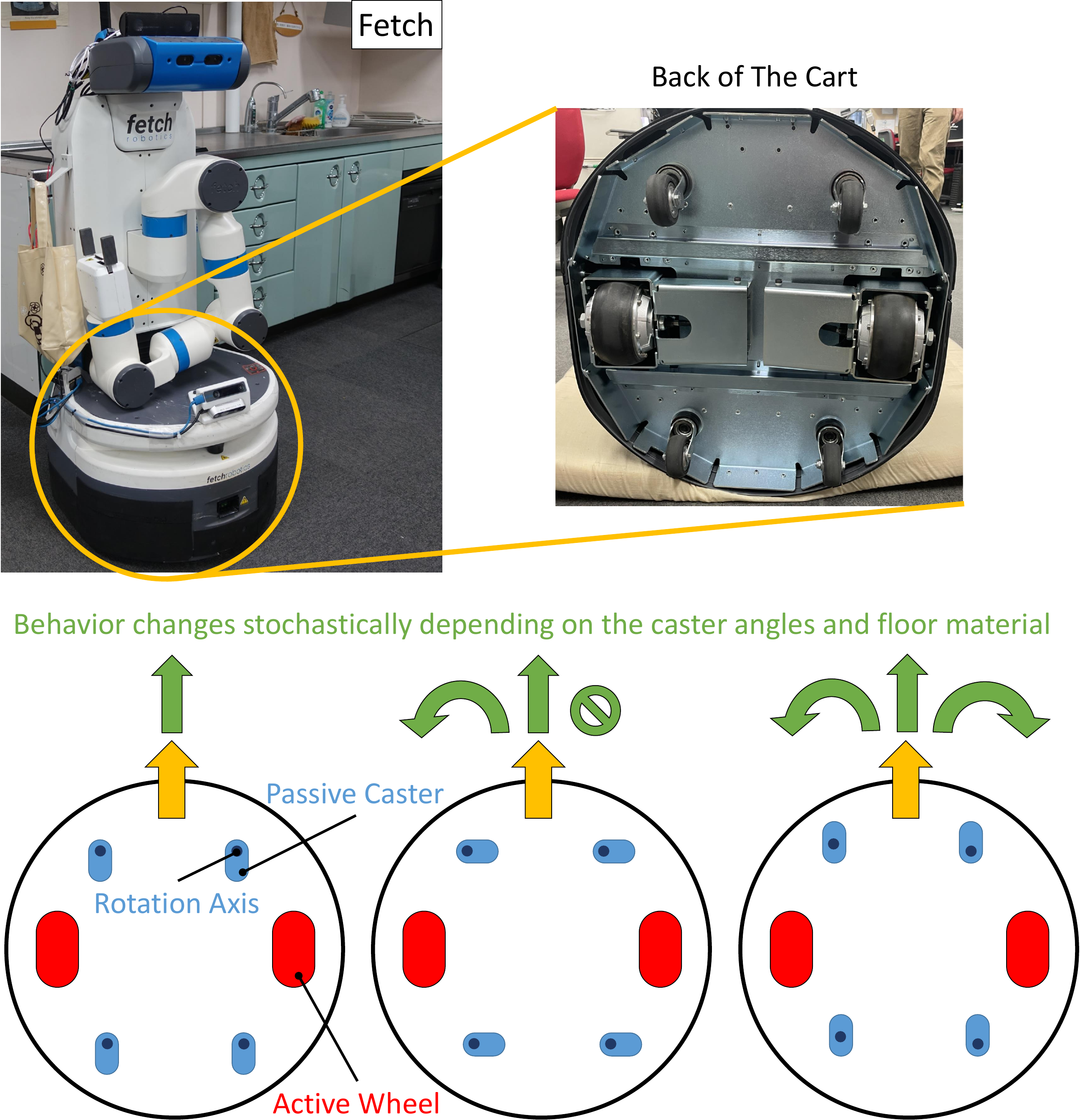}
  \vspace{-1.0ex}
  \caption{The motion of the robot Fetch is stochastic depending on the angles of the casters and the friction of the floor.}
  \label{figure:fetch-problem}
  \vspace{-1.0ex}
\end{figure}

\switchlanguage%
{%
  Various robot types have been developed so far, such as the mobile robot \cite{wise2016fetch}, the axis-driven humanoid \cite{hirukawa2004hrp}, the tendon-driven humanoid \cite{kawaharazuka2019musashi}, and the soft robot \cite{lee2017softrobotics}, etc.
  Among them, robots which are difficult to modelize due to the flexible bodies or partially observable states are challenging to handle with the conventional control methods.

  The problems are considered to be: (1) modeling is difficult to begin with, (2) the model can change online depending on the physical state of the robot and the surrounding environment, and (3) the model may have stochastic behavior.
  (1) is now widely known, and various deep learning approaches have been developed instead of the classical model-based controls.
  There are learning-based model predictive control \cite{lenz2015deepmpc, kawaharazuka2020regrasp}, reinforcement learning \cite{zhang2015reinforcement} and imitation learning \cite{zhang2018imitation}.
  As for (2), various approaches have been considered to solve the problem that control does not work as intended when the robot state or the surrounding environment deviates from that which was once learned.
  There is a method to obtain a control policy that works in any environment by performing actions in various environments \cite{lee2020anymal}, a method to sequentially incorporate changes in the body and environment into the network through online learning \cite{sofman2006online, kawaharazuka2020autoencoder}, and a method of implicitly embedding the information of the changes in the body and environment into a part of the network using parametric bias \cite{kawaharazuka2020dynamics}.
  As for (3), it may cause serious problems in some cases.
  Let us take an example of the mobile robot Fetch \cite{wise2016fetch} to illustrate the problem (\figref{figure:fetch-problem}).
  Fetch has two active wheels and four passive casters.
  As the angles of the casters depend on the previous movement, depending on the angles, it is often impossible to move backward.
  Also, the casters turn probabilistically which makes it likely to move in an unintended direction.
  These are characteristic of the older model of Fetch.
  In addition, these behaviors vary greatly depending on the floor material.
  Therefore, in order to solve these problems, it is necessary to include stochastic behaviors in the model while considering the changes in the model due to the surrounding environment such as the floor, and to minimize the variance of the predicted states to avoid unstable motion.
  As for the stochastic behavior, a method considering stochastic policy for reinforcement learning \cite{tedrake2004stochastic} and a method embedding the variability of human demonstrations into the network for imitation learning \cite{murata2013stochastic} have been developed.
  On the other hand, learning-based model predictive control that solves all of (1) -- (3) has not yet been developed.
}%
{%
  これまで台車型\cite{wise2016fetch}や軸駆動型ヒューマノイド\cite{hirukawa2004hrp}, 筋骨格ヒューマノイド\cite{kawaharazuka2019musashi}やSoft Robot \cite{lee2017softrobotics}など, 様々なロボットが開発されてきている.
  その中でも, 柔軟な身体を持つロボットや, 部分的にしか状態を観測できずモデル化が困難なロボットにおいては, これまでの古典的な制御で扱うことが難しい.

  これらの問題点は, (1)そもそもモデル化が困難であること, (2)そのモデルがロボットの身体状態や動作環境によって逐次的に変化してしまうこと, (3)モデルが確率的な挙動を示す場合があること, だと考える.
  (1)については現在広く周知されており, これまでの古典的なモデルベース制御ではなく, 深層学習型の様々なアプローチが開発されてきている.
  学習型モデル予測制御\cite{lenz2015deepmpc, kawaharazuka2020regrasp}や強化学習\cite{zhang2015reinforcement}・模倣学習\cite{zhang2018imitation}等がその例として挙げられる.
  (2)については, 一度学習した環境や身体の状態から外れたところでは意図したように制御が働かないという問題意識から, 様々なアプローチが成されている.
  様々な環境で動作を行うことでどのような環境においても動作する制御則を得る方法\cite{lee2020anymal}, オンライン学習により逐次的に身体や環境の変化をネットワークに取り込む方法\cite{sofman2006online, kawaharazuka2020autoencoder}, Parametric Biasを用いて暗黙的に身体や環境の情報をネットワークの一部に埋め込む方法\cite{kawaharazuka2020dynamics}等がある.
  (3)については, 場合によって深刻な問題を引き起こすことがある.
  台車型ロボットFetch \cite{wise2016fetch}を例に挙げてその問題を説明しよう.
  Fetch台車部分の駆動輪と受動輪の関係を\figref{figure:fetch-problem}に示す.
  この台車には2つの駆動輪と, 周りに4つのキャスターが配置されており, このキャスターが場合によって問題を引き起こす.
  キャスターの角度はその前の動きに依存し, その角度によっては後ろに進めなかったり, 確率的にキャスターが回り, 進めたとしても意図していない方向に大きく動いてしまうということが多い(古い型のFetchに特徴的である)
  また, この挙動は床の材質によっても大きく変化する.
  よって, この問題を解決するには, 床等の環境によるモデルの変化を考慮しつつ, モデルに平均と分散の確率的な挙動を入れ, 動作の際にはその分散を最小化しつつ行動することが必要になる.
  確率的挙動についてはこれまで, 強化学習の方策に確率的要素を入れた研究\cite{tedrake2004stochastic}, 人間のdemonstrationのばらつきをネットワークに埋め込み模倣学習に利用する手法等が開発されている\cite{murata2013stochastic}.
  一方, (1)(2)(3)全ての要素を解決する予測モデル型制御は開発されていない.
}%

\begin{figure*}[t]
  \centering
  \includegraphics[width=2.0\columnwidth]{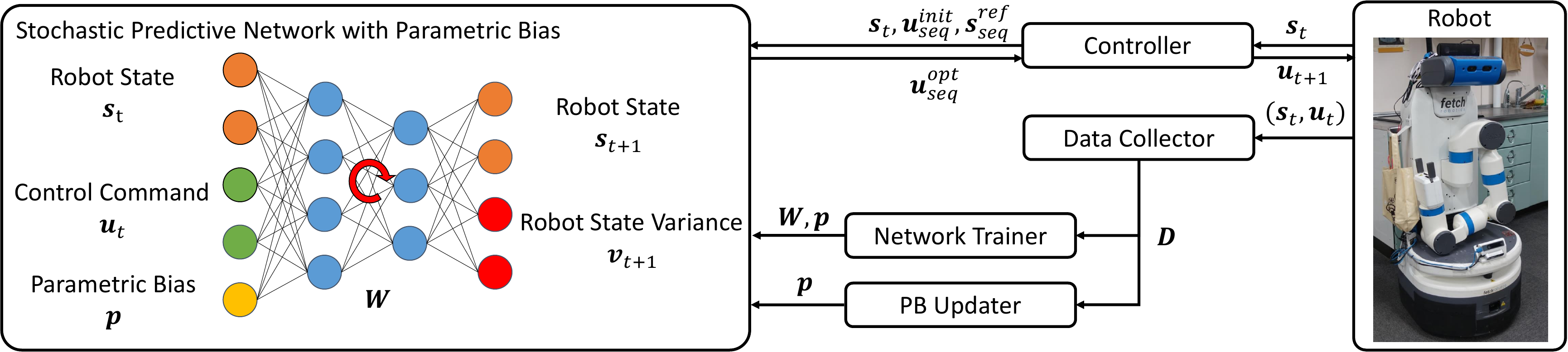}
  \vspace{-1.0ex}
  \caption{The overall system of using stochastic predictive network with parametric bias (SPNPB).}
  \label{figure:whole-system}
  \vspace{-1.0ex}
\end{figure*}

\switchlanguage%
{%
  In this study, we incorporate the ideas of \cite{kawaharazuka2020dynamics} and \cite{murata2013stochastic}, construct a predictive neural network that can consider stochastic behavior and implicitly embed environmental information as variables, and use this network for the adaptive robot control including variance minimization.
  The structure of this study is as follows.
  In \secref{sec:proposed-method}, we describe the construction of the proposed stochastic predictive network with parametric bias (SPNPB), its training, its environmental adaptation, and a control method including variance minimization.
  In \secref{sec:experiments}, we verify the effectiveness of our method on a mobile robot in simulation and the actual robot Fetch.
  The experimental results are discussed in \secref{sec:discussion}, and the conclusion is given in \secref{sec:conclusion}.
}%
{%
  そこで本研究では, \cite{kawaharazuka2020dynamics}や\cite{murata2013stochastic}の考え取り込み, 確率的挙動を考慮可能かつ暗黙的に環境情報を変数として埋め込み環境適応が可能な予測モデル型のニューラルネットワークの構築, これを用いた分散最小化を含む環境適応型ロボット制御の開発を行う.
  本研究の構成は以下のようになっている.
  \secref{sec:proposed-method}では本研究で提案するstochastic predictive network with parametric bias (SPNPB)の構築と, その学習, 環境適応, 分散最小化を含む制御手法について述べる.
  \secref{sec:experiments}では, 台車型ロボットのシミュレーション・mobile robot Fetchの実機において本手法の有効性を示す.
  \secref{sec:discussion}では実験結果について議論し, \secref{sec:conclusion}で結論を述べる.
}%

\section{Adaptive Control Including Variance Minimization Using Stochastic Predictive Network with Parametric Bias} \label{sec:proposed-method}
\switchlanguage%
{%
  We call the proposed network stochastic predictive network with parametric bias (SPNPB).
  The overall system of using SPNPB is shown in \figref{figure:whole-system}.
}%
{%
  本研究で提案するネットワークをstochastic predictive network with parametric bias (SPNPB)と呼ぶ.
  SPNPBとそれを取り巻く全体システムを\figref{figure:whole-system}に示す.
}%

\subsection{Stochastic Predictive Network with Parametric Bias} \label{subsec:spnpb}
\switchlanguage%
{%
  SPNPB can be expressed as follows,
  \begin{align}
    (\bm{s}_{t+1}, \bm{v}_{t+1}) = \bm{h}(\bm{s}_{t}, \bm{u}_{t}, \bm{p}) \label{eq:spnpb}
  \end{align}
  where $t$ is the current time step, $\bm{s}$ is the robot state such as the current image and joint velocity, $\bm{v}$ is the variance of $\bm{s}$ assuming a normal distribution, $\bm{u}$ is the control command, $\bm{p}$ is the parametric bias \cite{tani2002parametric}, and $\bm{h}$ is SPNPB.
  It differs from the conventional predictive model in that there is $\bm{p}$ for the input and $\bm{v}_{t+1}$ for the output.
  Note that the values of $\bm{s}$ and $\bm{u}$ are normalized using all the data obtained.
  Also, $\bm{v}_{t+1}$ represents the variance and is always positive, and so it is passed through exponential function.

  In this study, SPNPB has ten layers, which consist of four fully-connected layers, two LSTM layers \cite{hochreiter1997lstm}, and four fully-connected layers, in order.
  The number of units is set to \{$N_u+N_s+N_p$, 50, 20, 10, 10 (number of units in LSTM), 10 (number of units in LSTM), 10, 20, 50, $2N_s$\} (where $N_{\{u, s, p\}}$ is the dimensionality of $\{\bm{u}, \bm{s}, \bm{p}\}$).
  The activation function is hyperbolic tangent, and the update rule is Adam \cite{kingma2015adam}.
  The execution period of \equref{eq:spnpb} is set to 5 Hz.
}%
{%
  SPNPB数式で以下のように表せる.
  \begin{align}
    (\bm{s}_{t+1}, \bm{v}_{t+1}) = \bm{h}(\bm{s}_{t}, \bm{u}_{t}, \bm{p}) \label{eq:spnpb}
  \end{align}
  ここで, $t$は現在のタイムステップ, $\bm{s}$は画像や車輪速度等のロボット状態, $\bm{v}$はガウス分布を仮定した場合の$\bm{s}$の分散, $\bm{u}$は制御入力, $\bm{p}$はparametric bias \cite{tani2002parametric}, $\bm{h}$はSPNPBを表す.
  通常の予測モデルとは, $\bm{p}$が入力される点, $\bm{v}_{t+1}$が出力される点で異なっている.
  なお, $\bm{s}$, $\bm{u}$の値は得られた全データを使って正規化されている.
  また, $\bm{v}_{t+1}$は分散を表し常に正の値を取るため, exponentialを通して出力される.

  本研究においてSPNPBは10層とし, 順に4層のfully-connected layer, 2層のLSTM layer \cite{hochreiter1997lstm}, 4層のfully-connected layerからなる.
  ユニット数については, \{$N_u+N_s+N_p$, 50, 20, 10, 10 (LSTMのunit数), 10 (LSTMのunit数), 10, 20, 50, $2N_s$\}とした(なお, $N_{\{u, s, p\}}$は$\{\bm{u}, \bm{s}, \bm{p}\}$の次元数とする).
  activation functionはhyperbolic tangent, 更新則はAdam \cite{kingma2015adam}とした.
  また, \equref{eq:spnpb}の実行周期は5Hzとする.
}%

\subsection{Training of SPNPB} \label{subsec:training}
\switchlanguage%
{%
  We collect the data of $\bm{s}$ and $\bm{u}$ by random movements or human operation of the robot.
  By collecting data while changing the environment (e.g., the floor material or the physical state of the robot), we can embed them as implicit information into the parametric bias.
  For a trial $k$ executed in the same environment, the data $D_k=\{(\bm{s}^{k}_1, \bm{u}^{k}_{1}), (\bm{s}^{k}_2, \bm{u}^{k}_2), \cdots, (\bm{s}^{k}_{T_{k}}, \bm{u }^{k}_{T_{k}})\}$ ($1 \leq k \leq K$, where $K$ is the total number of trials and $T_{k}$ is the number of time steps for the trial $k$) is collected.
  Then, we obtain the data $D_{train}=\{(D_1, \bm{p}_1), (D_2, \bm{p}_2), \cdots, (D_{K}, \bm{p}_K)\}$ for training.
  The $\bm{p}_k$ is the parametric bias for the trial $k$, which is a variable that has a common value during that trial and a different value for different trials.
  We train SPNPB using this data $D_{train}$ and the following loss function through maximum likelihood estimation,
  \begin{align}
    P(s^{k}_{i, t}|D_{k, 1:t-1}, W, \bm{p}_{k}) &= \frac{1}{\sqrt{2\pi \hat{v}_{i, t}}}\exp{\left(-\frac{(\hat{s}^{k}_{i, t}-s^{k}_{i, t})^{2}}{2\hat{v}_{i, t}}\right)} \label{eq:prob-density}\\
    L_{likelihood}(W, \bm{p}_{1:K}|D_{train}) &= \prod^{K}_{k=1}\prod^{T_{k}}_{t=1}\prod^{N_{s}}_{i=1}P(s^{k}_{i, t}|D_{k, 1:t-1}, W, \bm{p}_{k}) \label{eq:likelihood}\\
    L_{train} &= -\log(L_{likelihood}) \label{eq:train-loss}
  \end{align}
  where $P$ is the probability density function, $\{s, v\}_{i}$ is the $\{s, v\}$ of the $i$-th sensor, $D_{k, 1:t-1}$ is the data of $D_k$ during $[1:t-1]$, $W$ is the weight of the current SPNPB, $\{\hat{s}, \hat{v}\}$ is the value of $\{s, v\}$ predicted from SPNPB using $D_{k, 1:t-1}$, $W$, and $\bm{p}_{k}$, and $\bm{p}_{1:K}$ is the vector of $\bm{p}_{k}$ for $1 \leq k \leq K$.
  We maximize $L_{likelihood}$ which represents the likelihood function for $W$ and $\bm{p}$ when given $D_{train}$.
  Here, we can simplify the calculation to the summation of $\log(P)$ by applying a transformation of \equref{eq:train-loss}, and minimize $L_{train}$.
  we usually update only the network weight $W$, but in this study, we update $\bm{p}_{k}$ at the same time.
  Note that all $\bm{p}_k$ are optimized with the initial value of $\bm{0}$.
}%
{%
  ランダムな動作や人間の操縦等によって, $\bm{s}$, $\bm{u}$に関するデータを取得する.
  この際, 環境(例えば床の材質やロボットの状態)を変化させながらデータを取得することで, それらを暗黙的な情報としてparametric biasに埋め込むことができる.
  ある同一の環境において取得した一連の試行$k$について, データ$D_k=\{(\bm{s}^{k}_1, \bm{u}^{k}_{1}), (\bm{s}^{k}_2, \bm{u}^{k}_2), \cdots, (\bm{s}^{k}_{T_{k}}, \bm{u}^{k}_{T_{k}})\}$を得る($1 \leq k \leq K$, $K$は全試行回数, $T_{k}$はその試行$k$に関する動作ステップ数とする).
  そして, 学習に用いるデータ$D_{train}=\{(D_1, \bm{p}_1), (D_2, \bm{p}_2), \cdots, (D_{K}, \bm{p}_K)\}$を得る.
  $\bm{p}_k$はその試行$k$に関するparametric biasであり, その試行中については共通の値で, 異なる試行については別の値となるような変数である.
  このデータ$D_{train}$と以下の損失関数を用いてSPNPBを学習させる.
  \begin{align}
    P(s^{k}_{i, t}|D_{k, 1:t-1}, W, \bm{p}_{k}) &= \frac{1}{\sqrt{2\pi \hat{v}_{i, t}}}\exp{\left(-\frac{(\hat{s}^{k}_{i, t}-s^{k}_{i, t})^{2}}{2\hat{v}_{i, t}}\right)} \label{eq:prob-density}\\
    L_{likelihood}(W, \bm{p}_{1:K}|D_{train}) &= \prod^{K}_{k=1}\prod^{T_{k}}_{t=1}\prod^{N_{s}}_{i=1}P(s^{k}_{i, t}|D_{k, 1:t-1}, W, \bm{p}_{k}) \label{eq:likelihood}\\
    L_{train} &= -\log(L_{likelihood}) \label{eq:train-loss}
  \end{align}
  ここで, $P$は確率密度関数, $\{s, v\}_{i}$は$i$番目のセンサの$\{s, v\}$, $D_{k, 1:t-1}$は$[1:t-1]$の間の$D_k$のデータ, $W$はSPNPBのネットワークの重み, $\{\hat{s}, \hat{v}\}$はデータ$D_k, 1:t-1$, 現在の重み$W$, 試行$k$に関する現在のparametric bias $\bm{p}_{k}$を使ってSPNPBから予測された$\{s, v\}$の値, $\bm{p}_{1:K}$は$1 \leq k \leq K$の$\bm{p}_{k}$をまとめたベクトルを表す.
  $L_{likelihood}$は $D_{train}$が与えられたときの$W$, $\bm{p}$に関する尤度関数を表し, これを最大化する問題を考える.
  このとき, \equref{eq:train-loss}のように変換をかませることで$\log(P)$の足し算へと計算を簡単にし, $L_{train}$の最小化問題に落としこむことができる.
  通常であればネットワークの重み$W$のみを更新していくが, 本研究では同時に$\bm{p}_{k}$も更新していくことになる.
  なお, 全$\bm{p}_k$は初期値を0として最適化される.
}%

\subsection{Online Update of Parametric Bias} \label{subsec:pb-update}
\switchlanguage%
{%
  By updating the parametric bias $\bm{p}$ online, the predictive model can adapt to the changing environment.
  While the robot is moving, the data $D_{update}$ of $\bm{s}$ and $\bm{u}$ are obtained at all times as in \secref{subsec:training}.
  Online learning of $\bm{p}$ starts when $N^{update}_{data}$, the number of data $D_{update}$, exceeds the threshold $N^{update}_{thre}$.
  The data exceeding $N^{update}_{max}$ is discarded from the oldest one.
  We use the same loss function as in \equref{eq:prob-density} -- \equref{eq:train-loss}, and update $\bm{p}$ with $N^{update}_{batch}$ batches and $N^{update}_{epoch}$ epochs.
  Here, the network weight $W$ is fixed and only $\bm{p}$ is updated.

  In this study, we set $N^{update}_{thre}=10$, $N^{update}_{max}=50$, $N^{update}_{batch}=N^{update}_{data}$ and $N^{update}_{epoch}=1$.
  Also, we use Momentum SGD \cite{qian1999momentum} for the update rule (Adam changes its behavior depending on $t$).
  The speed of the adaptation to environmental changes can be controlled by the learning rate of the update rule, but the larger the learning rate, the more unstable it becomes.
  The speed and accuracy of the adaptation can vary depending on the motion data.
}%
{%
  parametric bias $\bm{p}$をオンラインで更新し続けることで, 環境の変化に適応することができる.
  ロボットが動いている間に, 常に\secref{subsec:training}と同様な$\bm{s}$, $\bm{u}$のデータ$D_{update}$を取得しておく.
  この$D_{update}$のデータ数$N^{update}_{data}$が閾値$N^{update}_{thre}$を超えたところから$\bm{p}$のオンライン学習を始める.
  $N^{update}_{max}$を超えたデータは古いものから破棄する.
  損失関数は\equref{eq:prob-density} -- \equref{eq:train-loss}と同じものを使用し, バッチ数を$N^{update}_{batch}$, $N^{update}_{epoch}$として学習を行う.
  この際, ネットワークの重み$W$は固定し, $\bm{p}$のみを更新する.

  なお, 本研究では$N^{update}_{batch}=N^{update}_{data}$, $N^{update}_{thre}=10$, $N^{update}_{max}=50$, $N^{update}_{epoch}=1$, 更新則はMomentum SGD \cite{qian1999momentum}として学習を行う.
  環境変化への適応速度は更新則の学習率により調節ができるが, 大きいほど不安定になる.
  また, 動作データ次第でその速度や正確性は変わりうる.
}%

\subsection{Control Including Variance Minimization} \label{subsec:control}
\switchlanguage%
{%
  By using the trained SPNPB and $\bm{p}$ updated online to match the current environment, we can control the robot to execute the target task while reducing the variance of the motion.
  By reducing the variance, the robot can achieve the task while avoiding unstable motion.
  First, we determine the initial value $\bm{u}^{init}_{t:t+N_{seq}-1}$ of the control command $\bm{u}^{opt}_{t:t+N_{seq}-1}$ to be optimized.
  Here, $N_{seq}$ represents the length of the time series control command to be considered, and $\bm{u}^{\{init, opt\}}_{t:t+N_{seq}-1}$ is abbreviated as $\bm{u}^{\{init, opt\}}_{seq}$.
  The current state $\bm{s}_{t}$ is obtained and the following process is repeated to update $\bm{u}^{opt}_{seq}$,
  \begin{align}
    L_{control} &= ||\bm{s}^{ref}_{seq}-\hat{\bm{s}}_{seq}||_{2} + C_{variance}||\hat{\bm{v}}_{seq}||_{2}\label{eq:control-loss}\\
    \bm{u}^{opt}_{seq} &\gets \bm{u}^{opt}_{seq} - \gamma\partial L_{control}/\partial \bm{u}^{opt}_{seq}\label{eq:control-opt}
  \end{align}
  where $\bm{s}^{ref}_{seq}$ is the target state during $[t+1:t+N_{seq}]$, ${\{\hat{\bm{s}}, \hat{\bm{v}}\}}_{seq}$ is the value of $\{\bm{s}, \bm{v}\}$ predicted from the current SPNPB, $\bm{s}_{t}$, and $\bm{u}^{opt}_{seq}$, and $C_{variance}$ is the weight of the loss function.
  The first term on the right-hand side of \equref{eq:control-loss} is the loss to satisfy the given target state, and the second term is the loss to reduce the variance of the motion.
  \equref{eq:control-opt} updates $\bm{u}^{opt}_{seq}$ by using backpropagation technique and a gradient descent method.
  Here, $\bm{u}^{init}_{seq}$ is set as the value optimized in the previous time step $t-1$, $\bm{u}^{prev}_{t-1:t+N_{seq}-2}$, with $\bm{u}^{prev}_{t-1}$ removed and $\bm{u}^{prev}_{t+N_{seq}-2}$ duplicated.
  By using the value optimized in the previous step as the initial value, a faster convergence can be obtained.
  Although $\gamma$ can be a fixed value, in this study, we update $\bm{u}^{opt}_{seq}$ using $N^{control}_{batch}$ values of $\gamma$, which divide $[0:\gamma^{max}]$ logarithmically equally into $N^{control}_{batch}$ parts, at each iteration, and repeatedly use $\bm{u}^{opt}_{seq}$ with the lowest loss when running \equref{eq:control-loss} to obtain faster convergence.
  \equref{eq:control-loss} and \equref{eq:control-opt} are performed $N^{control}_{epoch}$ times at the current time step $t$, and the finally obtained $\bm{u}^{opt}_{seq}$ is sent to the actual robot.

  In this study, we set $N_{seq}=10$, $N^{control}_{batch}=10$, $N^{control}_{epoch}=3$, and $\gamma^{max}=3.0$.
  $C_{variance}$ is empirically set, and the larger it is, the more the variance can be taken into account, while the behavior tends to be unstable.
}%
{%
  学習されたSPNPB, 現在の環境に合わせて更新された$\bm{p}$から, 動きの分散を小さくしつつ指令したタスクを実行可能な制御を行うことができる.
  動きの分散が小さくなることで, 不安定な動きを回避しながら目的を達成できるようになる.
  まず, 最適化する制御入力$\bm{u}^{opt}_{t:t+N_{seq}-1}$の初期値$\bm{u}^{init}_{t:t+N_{seq}-1}$を決定する.
  ここで, $N_{seq}$は考慮する制御入力の時系列の長さを表し, 以降これらは$\bm{u}^{opt}_{seq}$, $\bm{u}^{init}_{seq}$のように省略する.
  現在状態$\bm{s}_{t}$を取得し, 以下の処理を繰り返すことで制御入力$\bm{u}^{opt}_{seq}$更新していく.
  \begin{align}
    L_{control} &= ||\bm{s}^{ref}_{seq}-\hat{\bm{s}}_{seq}||_{2} + C_{variance}||\hat{\bm{v}}_{seq}||_{2}\label{eq:control-loss}\\
    \bm{u}^{opt}_{seq} &\gets \bm{u}^{opt}_{seq} - \gamma\partial L_{control}/\partial \bm{u}^{opt}_{seq}\label{eq:control-opt}
  \end{align}
  ここで, $\bm{s}^{ref}_{seq}$は$[t+1:t+N_{seq}]$の間の指令状態, ${\{\hat{\bm{s}}, \hat{\bm{v}}\}}_{seq}$は$[t+1:t+N_{seq}]$の間における, 現在のSPNPB, $\bm{s}_{t}$, $\bm{u}^{opt}_{seq}$から予測された$\{\bm{s}, \bm{v}\}$の値, $C_{variance}$は損失関数の重みである.
  \equref{eq:control-loss}の右辺第一項は与えられた指令状態を満たすための損失, 右辺第二項は動作の分散を小さくするための損失である.
  \equref{eq:control-opt}は誤差逆伝播と最急降下法を用いて$\bm{u}^{opt}_{seq}$を更新する.
  このとき, $\bm{u}^{init}_{seq}$は, 前ステップ$t-1$で最適化された値$\bm{u}^{prev}_{t-1:t+N_{seq}-2}$について, $\bm{u}^{prev}_{t-1}$を削除し$\bm{u}^{prev}_{t+N_{seq}-2}$を複製したものとした.
  このように前ステップで最適化された値を初期値とすることで, より速い収束を得ることができる.
  また, $\gamma$は固定値でも良いが, 本研究では$[0:\gamma^{max}]$を対数的に等間隔に分けた$N^{control}_{batch}$個の$\gamma$を用いて$\bm{u}^{opt}_{seq}$をそれぞれ更新し, \equref{eq:control-loss}を実行した際に最も損失の小さかった$\bm{u}^{opt}_{seq}$を利用することを繰り返すことでより速い収束を得る.
  \equref{eq:control-loss} -- \equref{eq:control-opt}を現在のステップ$t$において$N^{control}_{epoch}$回行い, 最終的に得られた$\bm{u}^{opt}_{seq}$を実機へ送る.

  なお, 本研究では$N_{seq}=10$, $N^{control}_{batch}=10$, $N^{control}_{epoch}=3$, $\gamma^{max}=3.0$とした.
  $C_{variance}$は経験から設定され, 大きいほどより分散を考慮できる一方, 動作は不安定になりやすい.
}%

\begin{figure}[t]
  \centering
  \includegraphics[width=1.0\columnwidth]{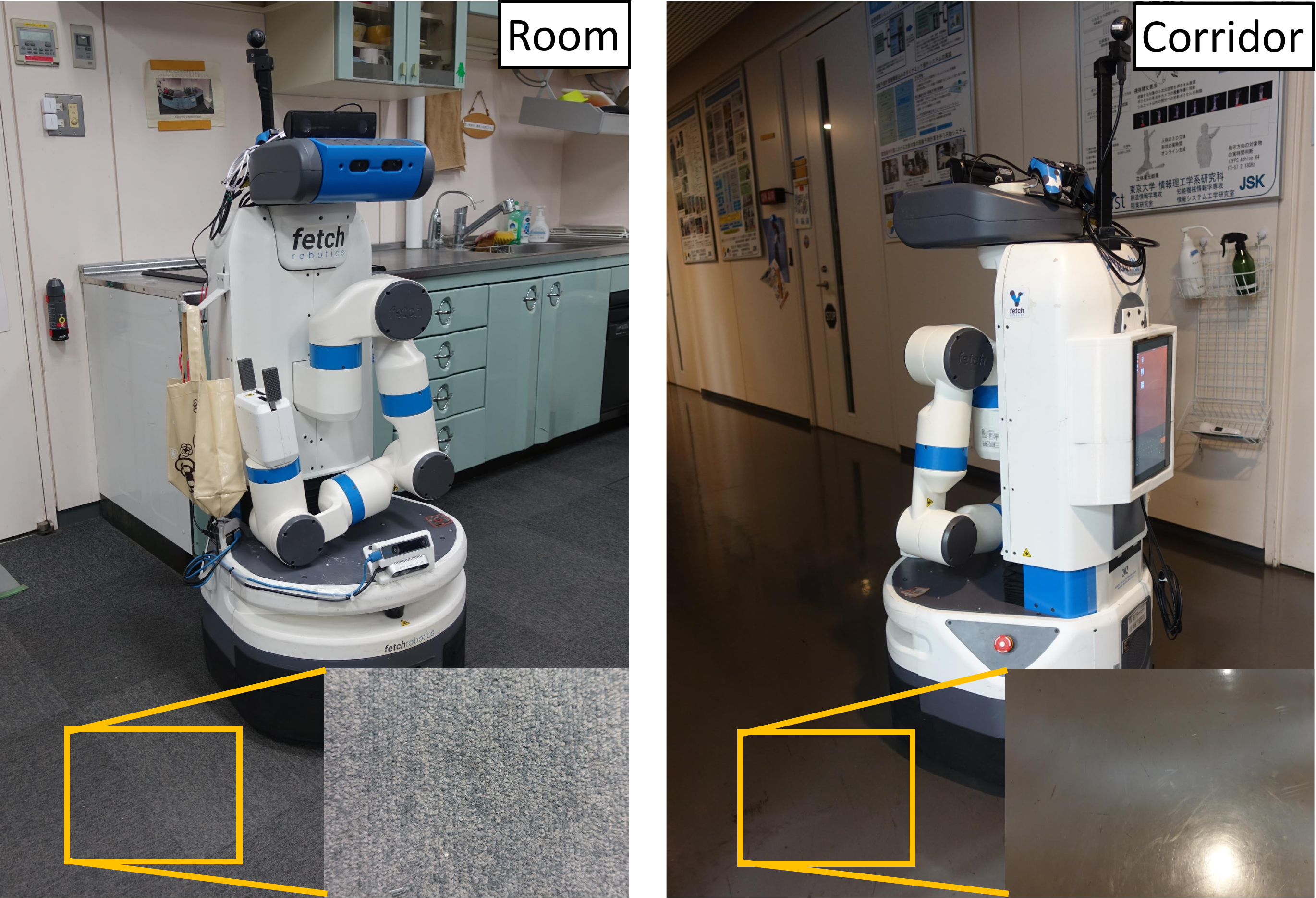}
  \vspace{-1.0ex}
  \caption{Two types of floor materials, Room and Corridor, for Fetch experiment.}
  \label{figure:experimental-setup}
  \vspace{-1.0ex}
\end{figure}

\section{Experiments} \label{sec:experiments}
\subsection{Experimental Setup} \label{subsec:experimental-setup}
\switchlanguage%
{%
  We perform two experiments using a simple mobile robot in simulation and the actual robot Fetch.
  In both experiments, $\bm{s}^{T}=\begin{pmatrix}w_{trans}&w_{rot}\end{pmatrix}$ and $\bm{u}^{T}=\begin{pmatrix}w^{ref}_{trans}&w^{ref}_{rot}\end{pmatrix}$, where $w_{\{trans, rot\}}$ is the measured velocity in the translational or rotational direction (the units are $\{[m/s], [rad/s]\}$) and $w^{ref}_{\{trans, rot\}}$ is the commanded velocity in the translational or rotational direction.
  It can also be expressed as $\bm{w}=\begin{pmatrix}w_{trans}&w_{rot}\end{pmatrix}^{T}$ and $\bm{w}^{ref}=\begin{pmatrix}w^{ref}_{trans}&w^{ref}_{rot}\end{pmatrix}^{T}$.
  We mainly optimize only for translational and rotational velocity, and do not consider trajectories in this study.
  In all experiments, $\bm{p}$ is set as two-dimensional value, and we do not reset the hidden states of LSTM through each experiment.
  The smaller $\bm{p}$ is, the easier the space of $\bm{p}$ is organized.
  In the actual robot experiment of this study, one dimension of $\bm{p}$ is sufficient because only two kinds of floors are used, but there was no problem even in two dimensions in this study.

  In the simulation experiment, we consider a simple system with a variance in the output motion depending on the motion speed as follows,
  \begin{align}
    w_{trans} &\gets w_{trans} + \alpha(w^{ref}_{trans}-w_{trans}) \nonumber\\
    &\;\;\;\; + \beta\textbf{N}(0, \frac{1}{|w_{trans}|+|w_{rot}|+0.1}) \label{eq:sim-trans}\\
    w_{rot} &\gets w_{rot} + \alpha(w^{ref}_{rot}-w_{rot}) + \beta\textbf{N}(0, 0.1) \label{eq:sim-rot}
  \end{align}
  where $\alpha$ is the coefficient for the rate of feedback, $\beta$ is the coefficient for the magnitude of the noise, and $\textbf{N}(\mu, \sigma)$ is the normally distributed noise with mean $\mu$ and standard deviation $\sigma$.
  While the noise on the rotation is small, the system has a large variance on the translation unless the velocity of the translation or rotation is large enough.
  The changes in $\alpha$ and $\beta$ corresponds to the changes in the robot state and the surrounding environment, such as motor characteristics and floor friction.

  In the Fetch experiment, we use the robot in two environments, Corridor and Room (\figref{figure:experimental-setup}).
  Corridor has a smooth and slippery floor, while Room has a carpeted floor with high friction.
  $w_{\{trans, rot\}}$ can be obtained from the visual odometry of the Intel Realsense T265 attached to the front of Fetch.
}%
{%
  台車のシミュレーション実験・Fetchを使った実機実験の2つを行う.
  どちらの実験においても, $\bm{s}^{T}=\begin{pmatrix}w_{trans}&w_{rot}\end{pmatrix}$, $\bm{u}^{T}=\begin{pmatrix}w^{ref}_{trans}&w^{ref}_{rot}\end{pmatrix}$とする(なお, $w_{\{trans, rot\}}$は並進・回転方向に測定された速度(単位は$\{[m/s], [rad/s]\}$), $w^{ref}_{\{trans, rot\}}$は並進・回転方向への指令速度を表す).
    $\bm{w}=\begin{pmatrix}w_{trans}&w_{rot}\end{pmatrix}^{T}$, $\bm{w}^{ref}=\begin{pmatrix}w^{ref}_{trans}&w^{ref}_{rot}\end{pmatrix}^{T}$と表現する場合もある.
  主に並進・回転速度に関する最適化のみ行い, 本研究では軌道については考えない.
  また, 全ての実験で$\bm{p}$は2次元とした.
  $\bm{p}$はできるだけ小さな方が組織化されやすい.
  本研究の実機実験では2種類の床しか用いないため1次元で十分であるが, 本研究では2次元でも問題はなかった.

  シミュレーションでは, 簡易的に以下のように制御入力に対して動作出力に分散が加わった系を考える.
  \begin{align}
    w_{trans} &\gets w_{trans} + \alpha(w^{ref}_{trans}-w_{trans}) \nonumber\\
    &\;\;\;\; + \beta\textbf{N}(0, \frac{1}{|w_{trans}|+|w_{rot}|+0.1}) \label{eq:sim-trans}\\
    w_{rot} &\gets w_{rot} + \alpha(w^{ref}_{rot}-w_{rot}) + \beta\textbf{N}(0, 0.1) \label{eq:sim-rot}
  \end{align}
  ここで, $\alpha$はフィードバックの割合を表す係数, $\beta$はノイズの大きさを表す係数, $\textbf{N}(\mu, \sigma)$は平均$\mu$, 分散$\sigma$の正規分布ノイズを表す.
  回転に関するノイズは小さい一方で, 並進に関しては並進または回転の速度がある程度大きくないと, 動作に大きな分散がのってしまう系になっている.
  $\alpha$, $\beta$の変化はモータの動きや床の摩擦等, ロボット身体や動作環境の状態の変化を簡易的に模擬している.

  Fetchを使った実験では, \figref{figure:experimental-setup}にあるCorridor, Roomの2つの環境で動作し, データを取得した.
  Corridorはツルツルと滑るような床なのに対して, Roomはカーペットが敷かれているため非常に摩擦が強い床になっている.
  $w_{\{trans, rot\}}$はFetchの台車前方についたIntel Realsense T265のVisual Odometryから得た.
}%

\begin{figure}[t]
  \centering
  \includegraphics[width=0.9\columnwidth]{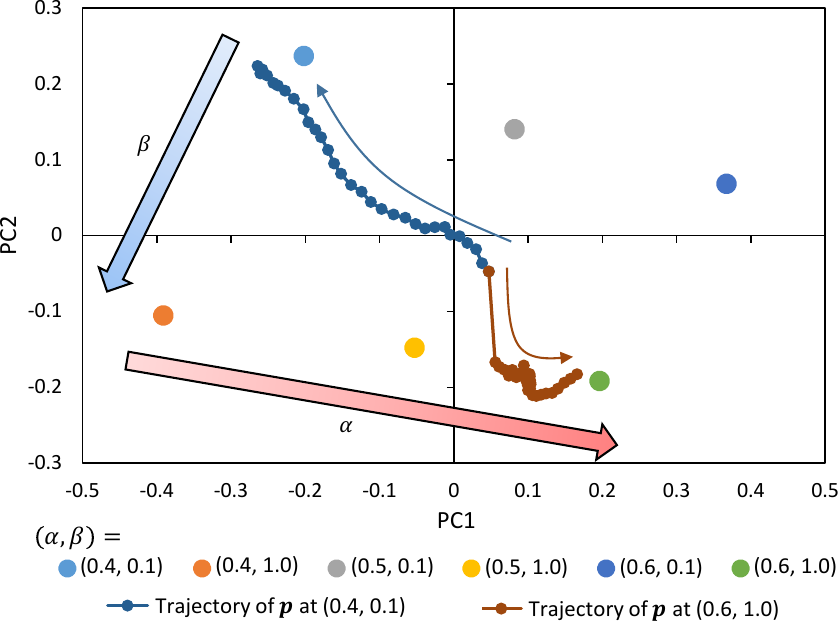}
  \vspace{-1.0ex}
  \caption{Training experiment in simulation: parametric biases $\bm{p}_{k}$ trained using the collected data and the trajectory of $\bm{p}$ updated by online learning.}
  \label{figure:sim-exp-pb}
  \vspace{-1.0ex}
\end{figure}

\begin{figure}[t]
  \centering
  \includegraphics[width=1.0\columnwidth]{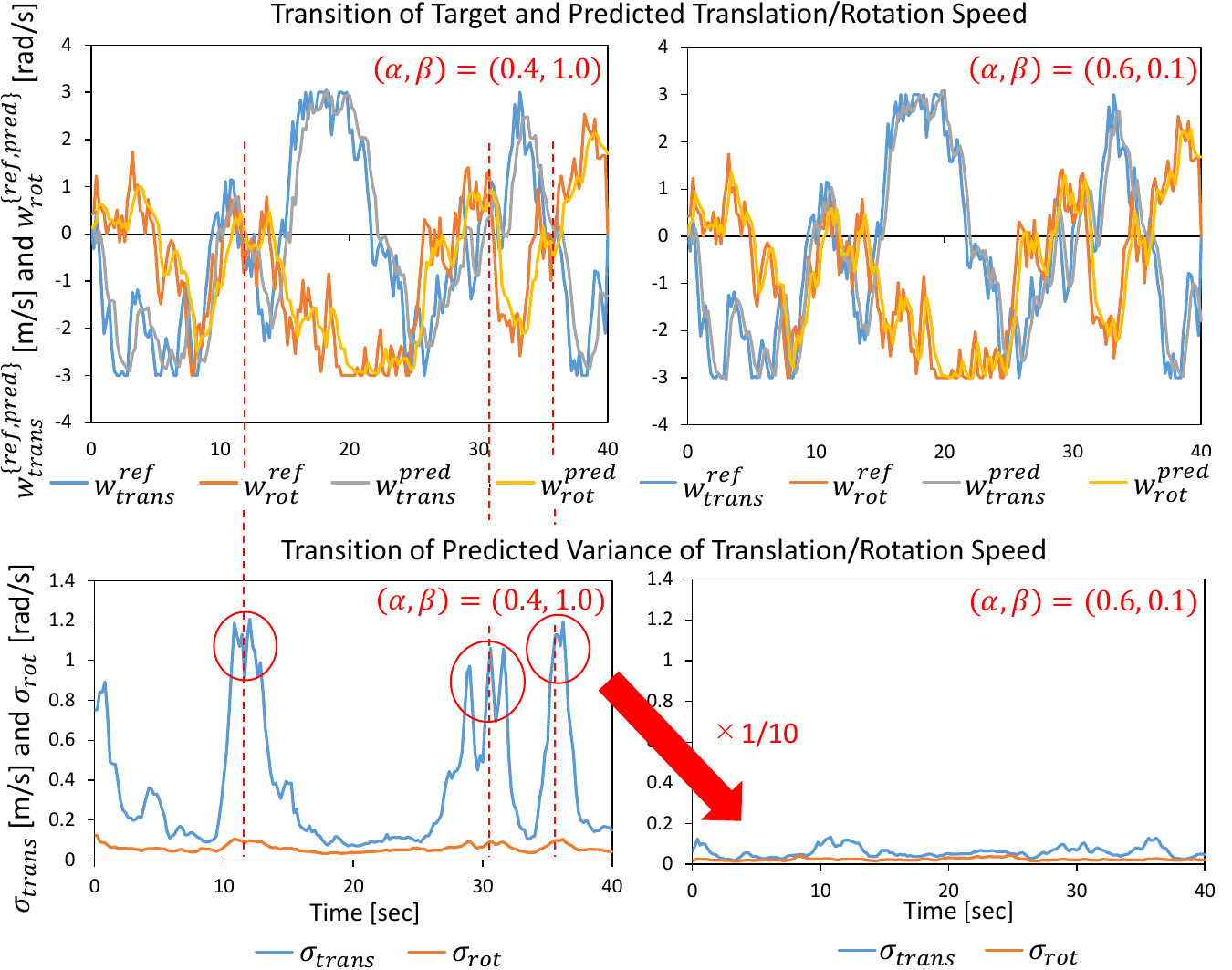}
  \vspace{-1.0ex}
  \caption{State estimation experiment in simulation: transition of the target velocity $\bm{w}^{ref}$, the predicted velocity $\bm{w}^{pred}$ from SPNPB, and the predicted standard deviation $\bm{\sigma}$ of $\bm{w}^{pred}$.}
  \label{figure:sim-exp-est}
  \vspace{-1.0ex}
\end{figure}

\begin{figure}[t]
  \centering
  \includegraphics[width=1.0\columnwidth]{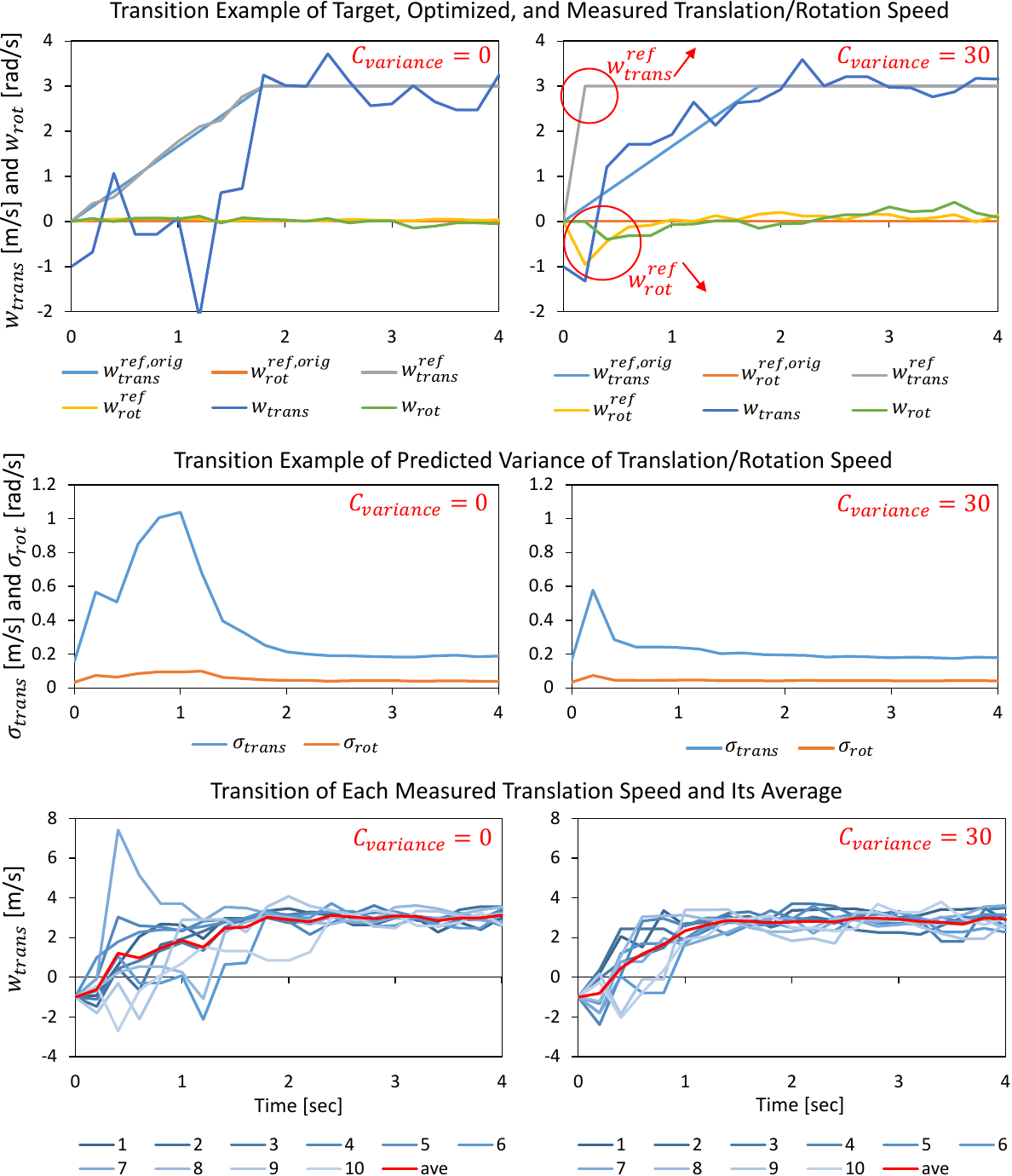}
  \vspace{-1.0ex}
  \caption{Control experiment in simulation: transition example of original target velocity $\bm{w}^{ref, orig}$, optimized velocity $\bm{w}^{ref}$, and measured velocity $\bm{w}$ in the upper graphs, transition of predicted standard deviation $\bm{\sigma}$ in the middle graphs, and ten transitions of $w_{trans}$ and its average in the lower graphs, when setting $C_{variance}=\{0, 30\}$.}
  \label{figure:sim-exp-control}
  \vspace{-1.0ex}
\end{figure}

\subsection{Simulation Experiment} \label{subsec:sim-exp}
\switchlanguage%
{%
  We run the simulator randomly with a control command such as $\bm{w}^{ref} \gets \max(-3, \min(\bm{w}^{ref} + \textrm{Random}(-1, 1), 3))$ (where $\textrm{Random}(a, b)$ returns a random value in the range of $[a, b]$).
  For each of the six combinations of $\alpha=\{0.4, 0.5, 0.6\}$ and $\beta=\{0.1, 1.0\}$, about 200 steps of data are obtained, and SPNPB is trained.
  The parametric bias $\bm{p}_{k}$ trained from the obtained data $D_{train}$ is represented in a two-dimensional plane through Principal Component Analysis (PCA) in \figref{figure:sim-exp-pb}.
  It can be seen that $\bm{p}_{k}$ is neatly arranged according to the sizes of $\alpha$ and $\beta$.

  Next, we describe the online learning of the parametric bias $\bm{p}$.
  The simulator is set to either $(\alpha, \beta)=(0.4, 0.1)$ or $(\alpha, \beta)=(0.6, 1.0)$, and $\bm{p}$ is updated online from the data when $\bm{w}^{ref}$ is randomly varied as described above.
  The transition of $\bm{p}$ is shown in \figref{figure:sim-exp-pb}.
  Note that the initial value of $\bm{p}$ is set to $\bm{0}$ ($\bm{p}=\bm{0}$ is not necessarily at the origin of the graph since it is transformed by PCA).
  As the online update progresses, $\bm{p}$ becomes closer to $\bm{p}_{k}$ trained with the parameters of the current simulation, indicating that the current state can be correctly recognized from the dynamics of the motion.

  Next, we examine the predicted mean $\bm{w}^{pred}$ of $\bm{w}$ and its variance $\bm{v}$ outputted from SPNPB with $\bm{w}^{ref}$ when moving randomly as described above.
  When $\bm{p}$ is set to $\bm{p}_k$ trained with $(\alpha, \beta)=(0.4, 1.0)$ or $(\alpha, \beta)=(0.6, 0.1)$, we show $w^{pred}_{\{trans, rot\}}$ and $\sigma_{\{trans, rot\}}$ ($\bm{\sigma}=\sqrt{\bm{v}}$) predicted from SPNPB in \figref{figure:sim-exp-est}.
  When $\alpha=0.6$, $\bm{w}^{pred}$ is closer to $\bm{w}^{ref}$ and converges faster than when $\alpha=0.4$, which can be expressed by SPNPB.
  Looking at $(\alpha, \beta)=(0.4, 1.0)$, when $|w^{pred}_{trans}|+|w^{pred}_{rot}|$ is small, $\sigma_{trans}$ is large (about 1.2 m/s), which is consistent with the characteristics of \equref{eq:sim-trans}.
  In addition, when $\beta=0.1$, the value of $\bm{\sigma}$ is about 1/10 of the value when $\beta=1.0$ each time, and the variance of the behavior is also correctly represented by SPNPB.

  Finally, we verify the control including variance minimization.
  The simulator is fixed to $(\alpha, \beta)=(0.5, 1.0)$, and $\bm{p}$ is also fixed to $\bm{p}_{k}$ obtained from the data at $(\alpha, \beta)=(0.5, 1.0)$.
  In this experiment, the target state $\bm{s}^{ref}$ at \secref{subsec:control} is $\bm{w}^{ref, orig}$, and the optimized control command $\bm{u}^{opt}$ sent to the robot is $\bm{w}^{ref}$.
  Here, we set $\bm{w}^{ref, orig}=\begin{pmatrix}3&0\end{pmatrix}^{T}$ when $\bm{w}=\begin{pmatrix}-1&0\end{pmatrix}^{T}$ (in addition, for the first two seconds, the target state is sent by linear interpolation from $\bm{w}^{ref, orig}=\begin{pmatrix}0&0\end{pmatrix}^{T}$), and verify the behavior of the robot state $\bm{w}$.
  Ten repetitions are conducted for $C_{variance}=0$ without considering the variance, and for $C_{variance}=30$ considering the variance.
  An example of the transitions of the original target state $\bm{w}^{ref, orig}$, the optimized command value $\bm{w}^{ref}$, and the measured robot state $\bm{w}$ is shown in the upper figure of \figref{figure:sim-exp-control}.
  $\bm{\sigma}$ predicted from SPNPB is also shown in the middle figure of \figref{figure:sim-exp-control}.
  Since variance is not taken into account when $C_{variance}=0$, the definition of $\bm{s}$ and $\bm{u}$ in this study results in $\bm{w}^{ref}=\bm{w}^{ref, orig}$.
  In this case, the value of $\sigma_{trans}$ increases to about 1.0, and $w_{trans}$ does not follow $w^{ref}_{trans}$ well, indicating that the motion is unstable.
  On the other hand, when $C_{variance}=30$, $\bm{w}^{ref}$ does not follow $\bm{w}^{ref, orig}$, $w^{ref}_{trans}$ rises sharply, and $w^{ref}_{rot}$ is not constant at 0 but changes in the early stage of motion.
  By this control command, $w_{trans}$ changes relatively stably, avoiding the unstable state in which $|w_{trans}| + |w_{rot}|$ is close to 0.
  The ten transitions of $w_{trans}$ and their average are shown in the lower figure of \figref{figure:sim-exp-control}.
  As a whole, when $C_{variance}=0$, there are some unstable behaviors in the early stage of the motion, but when $C_{variance}=30$, they can be avoided.
  On the other hand, when $C_{variance}=0$, the average of $w_{trans}$ follows $w^{ref, orig}_{trans}$ well, while when $C_{variance}=30$, $w^{ref, orig}_{trans}=3$ is reached faster than the original target state transition, which can be seen as sacrificing the accuracy of the task in order to avoid unstable behavior.
}%
{%
  シミュレータを$\bm{w}^{ref} \gets \max(-3, \min(\bm{w}^{ref} + \textrm{Random}(-1, 1), 3))$のような制御入力でランダムで動かす(なお, $\textrm{Random}(a, b)$は$[a, b]$の範囲内のランダムな値を返す関数とする).
  $\alpha=\{0.4, 0.5, 0.6\}$, $\beta=\{0.1, 1.0\}$の6つの組み合わせについてそれぞれ約200ステップのデータを取得する.
  得られたデータ$D_{train}$からSPNPBを学習した際のparametric bias $\bm{p}_{k}$をPrincipal Component Analysis (PCA)に通し2次元平面に表現した結果を\figref{figure:sim-exp-pb}に示す.
  $\alpha$, $\beta$の大きさに応じて$\bm{p}_{k}$が綺麗に配置されていることがわかる.

  次に, parametric bias $\bm{p}$のオンライン学習について述べる.
  シミュレータを$(\alpha, \beta)=(0.4, 0.1)$または$(\alpha, \beta)=(0.6, 1.0)$のニ種類に設定し, 上述のようにランダムに$\bm{w}^{ref}$を変化させた際のデータからオンラインで$\bm{p}$を更新したときの$\bm{p}$の挙動を\figref{figure:sim-exp-pb}に示す.
  なお, 初期は$\bm{p}=\bm{0}$としている(PCAにより変換されるため, $\bm{p}=\bm{0}$が原点にあるとは限らない).
  学習が進むごとに$\bm{p}$は, 訓練時に得られた$\bm{p}_{k}$の中でも現在のシミュレーションのパラメータの$\bm{p}_{k}$に近づいていき, 動きのダイナミクスから現在の状態を把握できていることがわかる.

  次に, 上述のようにランダムに動いた際の$\bm{w}^{ref}$から, SPNPBによってどのような$\bm{w}$の予測平均$\bm{w}^{pred}$とその分散$\bm{v}$が出力されるかを検証する.
  $\bm{p}$を$(\alpha, \beta)=(0.4, 1.0)$のデータから得られた$\bm{p}_k$, $(\alpha, \beta)=(0.6, 0.1)$のデータから得られた$\bm{p}_k$のニ種類に設定した際に, SPNPBから予測される$w^{pred}_{\{trans, rot\}}$, $\sigma_{\{trans, rot\}}$ ($\bm{\sigma}=\sqrt{\bm{v}}$)の推移を\figref{figure:sim-exp-est}に示す.
  $\alpha=0.6$のときは$\alpha=0.4$のときに比べ, $\bm{w}^{pred}$がより$\bm{w}^{ref}$に近く, 速く収束していることがSPNPBにより表現できている.
  $(\alpha, \beta)=(0.4, 1.0)$のときを見ると, $|w^{pred}_{trans}|+|w^{pred}_{rot}|$が小さいときに$\sigma_{trans}$は大きくなっており, \equref{eq:sim-trans}の特性と一致している.
  また, $\beta=0.1$のときは$\beta=1.0$のときに比べて$\bm{\sigma}$の値が約1/10になっており, 動作の分散についてもSPNPBにより正しく表現できている.

  最後に, 分散最小化を含む制御について検証する.
  シミュレータを$(\alpha, \beta)=(0.5, 1.0)$に固定, $\bm{p}$も$(\alpha, \beta)=(0.5, 1.0)$のデータから得られた$\bm{p}_{k}$に固定する.
  本実験では\secref{subsec:control}における指令値$\bm{s}^{ref}$を$\bm{w}^{ref, orig}$, 最適化されて最終的にロボットに送られる指令値$\bm{u}^{opt}$を$\bm{w}^{ref}$とする.
  ここで, ロボットが$\bm{w}=\begin{pmatrix}-1&0\end{pmatrix}^{T}$の状態から指令値として$\bm{w}^{ref, orig}=\begin{pmatrix}3&0\end{pmatrix}^{T}$を送る場合(なお, 最初の2秒間は$\bm{w}^{ref, orig}=\begin{pmatrix}0&0\end{pmatrix}^{T}$から線形補間して指令値を送る)について, \secref{subsec:control}の制御を行い, ロボットの状態$\bm{w}$の挙動について確認する.
  なお, 分散を考慮しない$C_{variance}=0$, 分散を考慮する$C_{variance}=30$の2つについてそれぞれ10回実験して検証している.
  元の指令値$\bm{w}^{ref, orig}$, 最適化された指令値$\bm{w}^{ref}$, 得られたロボット状態$\bm{w}$の遷移の一例を\figref{figure:sim-exp-control}の上図に示す.
  また, その際にSPNPBから予測される$\bm{\sigma}$を\figref{figure:sim-exp-control}の中図に示す.
  $C_{variance}=0$の際は, 一切分散が考慮されないため, 本研究の$\bm{s}$, $\bm{u}$の定義だと, $\bm{w}^{ref}=\bm{w}^{ref, orig}$となっていることがわかる.
  この場合, $\sigma_{trans}$の値が約1.0まで上がり, $w_{trans}$がうまく$w^{ref}_{trans}$に追従せず, 動作が不安定になっていることがわかる.
  これに対して, $C_{variance}=30$の際は, $\bm{w}^{ref}$は$\bm{w}^{ref, orig}$には追従せず, $w^{ref}_{trans}$が一気に大きく上昇し, $w^{ref}_{rot}$も0で一定ではなく動作初期に変化していることがわかる.
  これにより, $|w_{trans}| + |w_{rot}|$が0に近い不安定な状態を回避し, 比較的安定して$w_{trans}$が変化していることがわかる.
  この動作を10回行った際の$w_{trans}$の遷移とその平均を\figref{figure:sim-exp-control}の下図に示す.
  全体として, $C_{variance}=0$のときは動作初期に不安定な動作が散見されるが, $C_{variance}=30$のときはそれを回避できていると言える.
  一方, 平均を取ると$C_{variance}=0$のときは$w_{trans}$が$w^{ref, orig}_{trans}$に良く追従しており, $C_{variance}=30$のときは指令値遷移に対してより速く$w^{ref, orig}_{trans}=3$に到達してしまっており, 不安定な動作を回避する代わりに精度を犠牲にしているとも取れる.
}%

\begin{figure}[t]
  \centering
  \includegraphics[width=0.8\columnwidth]{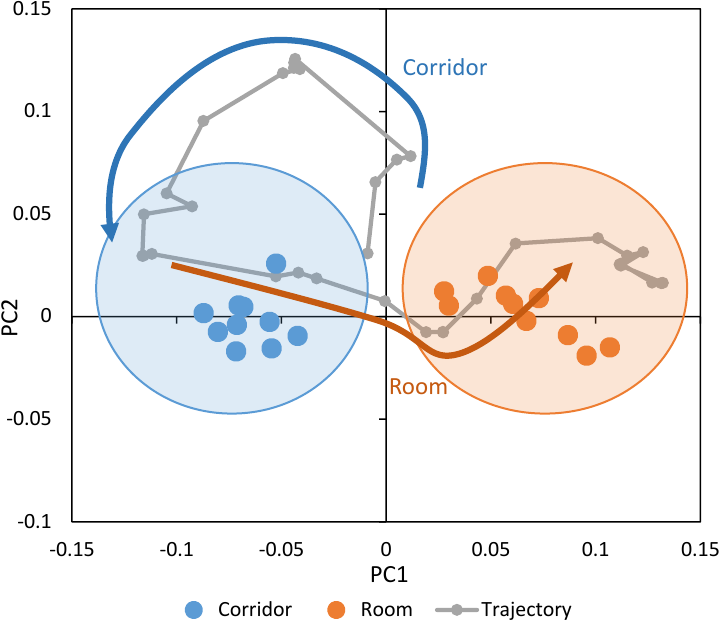}
  \vspace{-1.0ex}
  \caption{Training experiment in Fetch: parametric biases $\bm{p}_{k}$ trained using the collected data and the trajectory of $\bm{p}$ updated by online learning.}
  \label{figure:fetch-exp-pb}
  \vspace{-1.0ex}
\end{figure}

\begin{figure*}[t]
  \centering
  \includegraphics[width=2.0\columnwidth]{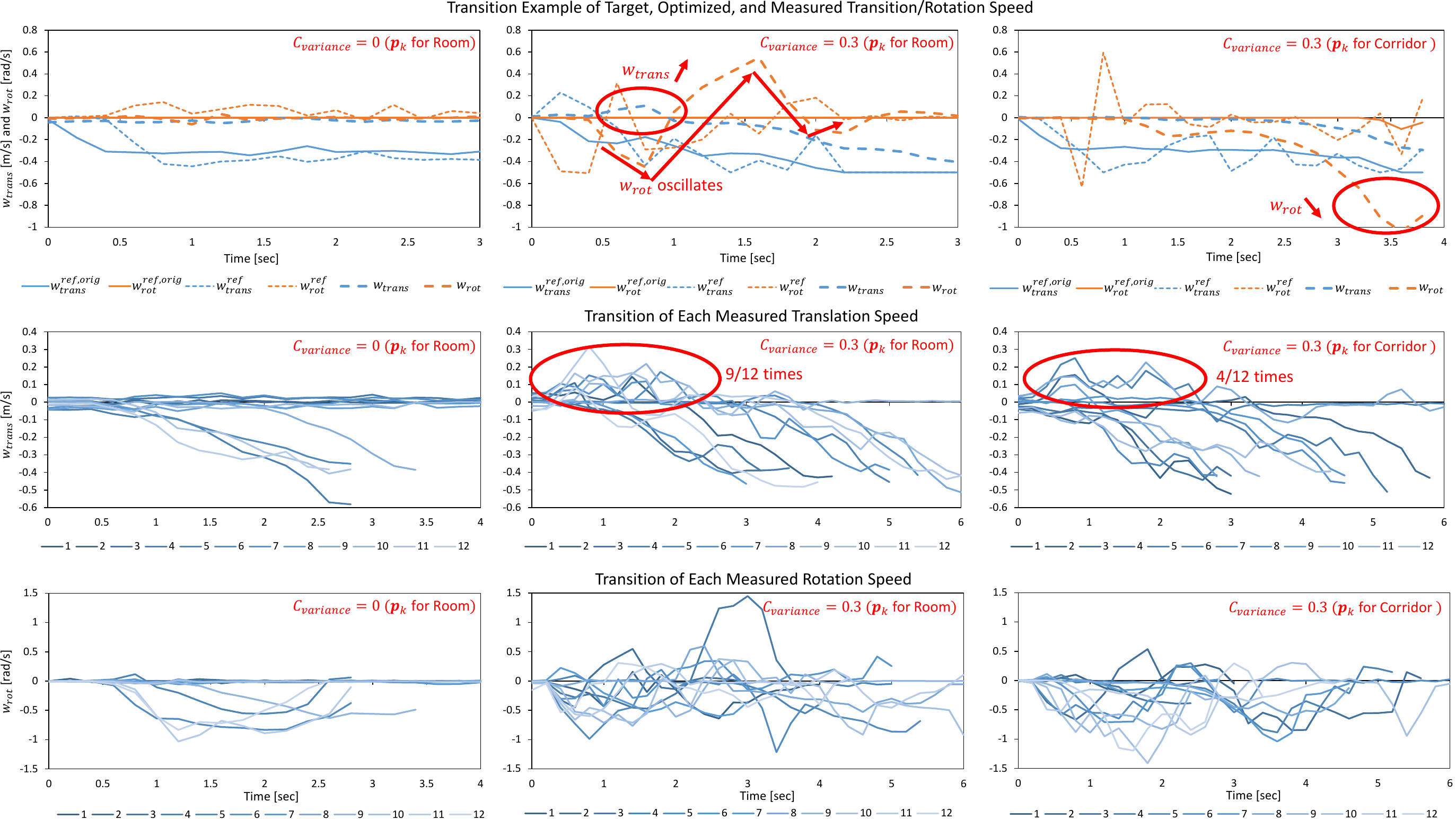}
  \vspace{-1.0ex}
  \caption{Control experiment in Fetch: transition example of original target velocity $\bm{w}^{ref, orig}$, optimized velocity $\bm{w}^{ref}$, and measured velocity $\bm{w}$ in the upper graphs, and twelve transitions of $w_{\{trans, rot\}}$ in the middle and lower graphs, when setting $C_{variance}=0$ ($\bm{p}_k$ for Room), $C_{variance}=0.3$ ($\bm{p}_k$ for Room), and $C_{variance}=0.3$ ($\bm{p}_k$ for Corridor).}
  \label{figure:fetch-exp-control}
  \vspace{-1.0ex}
\end{figure*}

\subsection{Fetch Experiment} \label{subsec:fetch-exp}
\switchlanguage%
{%
  The data $D_{train}$ is collected by performing various motions in two environments of Corridor and Room with a joystick controller.
  The parametric bias $\bm{p}_{k}$ of SPNPB trained from the obtained data $D_{train}$ is represented on a two-dimensional plane through PCA as shown in \figref{figure:fetch-exp-pb}.
  We can see that $\bm{p}_{k}$ is neatly divided into left and right for Corridor and Room.
  The transition of $\bm{p}$ by online update is shown in \figref{figure:fetch-exp-pb} when the robot is operated with a joystick from Corridor to Room.
  Note that the initial value of $\bm{p}$ is set to $\bm{0}$.
  It can be seen that $\bm{p}$ initially moves around $\bm{p}_{k}$ for Corridor and then gradually moves around $\bm{p}_{k}$ for Room as the location of the robot is moved to Room.

  We will examine the motion of moving backward after rotation, which is the challenging motion that follows the target state the least among the motions we have performed so far.
  When the robot rotates, the caster becomes perpendicular to the active wheel, and when the robot moves backward in this state, the motion often gets stuck.
  Note that it is possible for the robot to move forward easily even after rotation due to several unknown reasons, possibly related to the suspension placement, direction-dependent motor output, or load distribution.
  This experiment is conducted in Room, and we compare three cases: $C_{variance}=\{0, 0.3\}$ when using $\bm{p}$ updated in Room, and $C_{variance}=0.3$ when using $\bm{p}$ updated in Corridor (the behaviors with $C_{variance}=0$ when using $\bm{p}$ for Room and Corridor are almost the same).
  In this experiment, we added the term $||\bm{w}^{ref, orig}-\bm{w}^{ref}||_{2}$ for \equref{eq:control-loss} to make the motion more stable.
  Also, considering that the larger $\bm{w}$ is, the larger the variance $\bm{v}$ will inevitably be, we rewrite the second term on the right-hand side of \equref{eq:control-loss} as $||\hat{\bm{v}}_{seq}\oslash\bm{\bm{s}}_{seq}||_{2}$ ($\oslash$ expresses element-wise division).
  After one rotation (unified to the right rotation; if rotated to the left, the sign of $w_{rot}$ is reversed), we send a target state of -1 m/s in the translational direction, but the actual target state $\bm{w}^{ref, orig}$ gradually increases according to the current $\bm{w}$ due to the safety mechanism.
  The transitions of $\bm{w}^{ref, orig}$, $\bm{w}^{ref}$ and $\bm{w}$ are shown in the upper figure of \figref{figure:fetch-exp-control} as an example of a typical motion under each condition.
  It can be seen that when $C_{variance}=0$, there is no significant difference between $\bm{w}^{ref, orig}$ and $\bm{w}^{ref}$, but $w_{trans}$ remains 0 and the motion is completely stuck.
  On the other hand, when $C_{variance}=0.3$ and $\bm{p}$ for Room is used, $\bm{w}^{ref, orig}$ and $\bm{w}^{ref}$ differ significantly, and $w_{trans}$ begins to move after about 1.5 seconds.
  A characteristic of this condition is that $w_{trans}$ first moves forward to change the direction of the caster, and $\bm{w}_{rot}$ moves oscillatorily to prevent the robot from getting stuck.
  In the case where $C_{variance}=0.3$ and $\bm{p}$ for Corridor is used, $w_{trans}$ can also move.
  In this condition, $w_{trans}$ does not move forward but directly moves backward more often than in the case using $\bm{p}$ for Room.
  In addition, there are many cases where $w_{rot}$ moves significantly in the same direction as the previous rotation, that is, it moves backward while rotating.
  $w_{\{trans, rot\}}$ for the twelve repetitions of this experiment in each condition are shown in the middle and bottom figures of \figref{figure:fetch-exp-control}.
  Looking at $w_{trans}$, we can see that when $C_{variance}=0$, the motion is stuck in more than half of the trials, while when $C_{variance}=0.3$, the motion in the backward direction succeeds in all but one or two trials.
  When $C_{variance}=0.3$, the numbers of cases of moving backward after moving forward once were 9/12 times when using $\bm{p}$ for Room and 4/12 times when using $\bm{p}$ for Corridor.
  Looking at $w_{rot}$, we can see that $\bm{w}_{rot}$ has basically moved in the same direction as the rotation before backward motion ($w_{rot}<0$ because of the right rotation).
  On the other hand, when $C_{variance}=0.3$ and $\bm{p}$ for Room are used, $w_{rot}$ moves as if it was oscillating around 0 and the robot moves relatively straight.
}%
{%
  人間がjoystickを使って様々な動きをさせてデータ$D_{train}$を取得する.
  得られたデータ$D_{train}$からSPNPBを学習した際のparametric bias $\bm{p}_{k}$をPCAを通して2次元平面に表現した結果を\figref{figure:fetch-exp-pb}に示す.
  RoomとCorridorで$\bm{p}_{k}$が綺麗に左と右に分けられていることがわかる.
  また, $\bm{p}=\bm{0}$の状態から, Corridor, Roomの順にjoystickを使って動作させたときの, オンライン学習による$\bm{p}$の遷移を\figref{figure:fetch-exp-pb}に示す.
  $\bm{p}$は始めCorridorにおいて得られた$\bm{p}_{k}$の周辺に動き, その後Roomに場所を移動すると$\bm{p}$も徐々にRoomにおいて得られた$\bm{p}_{k}$周辺に移動していくことがわかる.

  これまで動作させた中でもっとも指令値通りに動かない動作である, ロボットを一回転させた後に後ろに進むという動作について検証を行う.
  一回転させるとキャスターが能動輪に対して垂直に向き, その状態で後ろに進むと, 動作がスタックしてしまうことが多い.
  なお, 前進についてはサスペンションの関係か容易に進むことができる.
  本実験はRoomで行い, Roomにおいて更新した$\bm{p}$を使う場合について$C_{variance}=\{0, 0.3\}$を, Corridorにおいて更新した$\bm{p}$を使う場合について$C_{variance}=0.3$の場合の三種類について比較する.
  本実験ではより安定的に動作させるために, \equref{eq:control-loss}について, $||\bm{w}^{ref, orig}-\bm{w}^{ref}||_{2}$という項を足している.
  また, $\bm{w}$が大きいほど必然的に分散$\bm{v}$は大きくなると考え, \equref{eq:control-loss}の右辺第二項を$||\hat{\bm{v}}_{seq}\oslash\hat{\bm{s}}_{seq}||_{2}$のように書き換えている.
  一回転(右回転に統一する)した後, 並進方向について-1 m/sを指令値として送るが, 安全装置により現在の$\bm{w}$に応じて徐々に指令値$\bm{w}^{ref, orig}$は上がっていく.
  それぞれの条件における, 代表的な動作の一例について$\bm{w}^{ref, orig}$, $\bm{w}^{ref}$, $\bm{w}$の遷移を\figref{figure:fetch-exp-control}の上図に示す.
  $C_{variance}=0$の場合には, $\bm{w}^{ref, orig}$と$\bm{w}^{ref}$の間に大きな違いはないが, $w_{trans}$は0のままで完全に動作がスタックしていることがわかる.
  それに対して, $C_{variance}=0.3$かつRoomの$\bm{p}$を使った場合は$\bm{w}^{ref, orig}$と$\bm{w}^{ref}$が大きく異なり, $w_{trans}$も1.5秒後程度から動き始めていることがわかる.
  この条件に特徴的なのは, $w_{trans}$が最初前方に動くことでキャスターの向きを変化させること, そして$\bm{w}_{rot}$を振動させることでスタックしないように動かしている点である.
  $C_{variance}=0.3$かつCorridorの$\bm{p}$を使った場合も同様に$w_{trans}$を動作させることができている.
  この条件では, Roomの$\bm{p}$を使った場合に比べ, $w_{trans}$を前方に動かさず, 直接後方に対して指令値を送る動作が多い.
  また, $w_{rot}$が前の回転時の動作と同じ方向に大きく動く, つまり回転しながら後方に下がる場合が多い.
  12回の試行について$w_{\{trans, rot\}}$を\figref{figure:fetch-exp-control}の中図・下図に示す.
  $w_{trans}$を見ると, $C_{variance}=0$の場合は半分以上の試行で動作がスタックしているのに対して, $C_{variance}=0.3$の場合は1,2回の試行を除いて後進方向への動作に成功している.
  なお, $C_{variance}=0.3$で一度前進してから後進するケースはRoomの$\bm{p}$を使った場合は9/12回, Corridorの$\bm{p}$を使った場合は4/12回であった.
  $w_{rot}$を見ると, 基本的に後進前の回転(右回転のため$w_{rot}<0$)と同じ方向に$\bm{w}_{rot}$が動いてしまっていることがわかる.
  これに対して, $C_{variance}=0.3$かつRoomの$\bm{p}$を使った場合は$w_{rot}$が0付近で振動したような挙動になっており, 比較的真っ直ぐ進む.
}%

\section{Discussion} \label{sec:discussion}
\switchlanguage%
{%
  We discuss the experimental results of this study.
  First, simulation experiments show that for a simple system with stochastic behavior, the dynamics of the system can be neatly self-organized in the network by parametric bias from the obtained motion data without explicitly providing the parameters of the system.
  In addition, the current state of the system can be recognized by updating the parametric bias online from the motion data.
  It is found that $\bm{p}$ contains the characteristics of the system, and the inference by SPNPB outputs the expected behavioral transitions and variances regarding the system parameters $\alpha$ and $\beta$.
  Using the trained SPNPB, the learning control including variance minimization can realize the task while avoiding unstable behaviors with high variance.
  Next, the experiments on Fetch showed that the implicit parameters such as friction and texture of the floor can be self-organized in the network by parametric bias even on the actual robot.
  The system can detect the change of the floor material by online learning, and can recognize the environment in which it moves.
  Finally, we applied our method to backward motion after rotation, in which the robot often gets stuck, and found that the intended backward motion can be generated by optimization.
  In contrast to direct backward motion, a stuck-free motion was generated by moving forward once, changing the direction of the casters, and continuing to move oscillatorily in the direction of rotation.
  In addition, the behavior changed when $\bm{p}$ was changed.
  When $\bm{p}$ for Corridor was used, there were many behaviors that forced the robot to keep moving backward while rotating, without moving forward.
  This is considered to be due to the fact that Corridor is a floor with low friction, which makes it unlikely to get stuck even if it does not move forward.
  In this experiment, the use of inappropriate $\bm{p}$ did not make a significant difference in performance, but it may have an effect on other motions.

  We think that there are two major limitations in this study.
  The first one is that we cannot set the execution period high at present, because it takes time for optimization.
  Normally, the response is instantaneous because $w^{ref}=w^{ref, orig}$, but in this study, the response is delayed by the time taken to optimize $w^{ref}$ (200 msec here).
  Therefore, it is considered to create another network that outputs the optimized $w^{ref}$ in a single forwarding.
  Second, the stochastic behavior is modelized as a normal distribution.
  Although this study was effective in the experimental case of the mobile robot, it is actually not strictly approximated by a normal distribution.
  When we apply this method to more complex systems, we may not be able to modelize it well.
  It is necessary to apply this study to various systems and investigate its effective range in the future.

  In this study, we used a mobile robot as an example, but this study is expected to be applied to musculoskeletal humanoids and soft robots, which are more complex and difficult to modelize.
  We would like to use this study to modelize the online change in dynamics and stochastic behaviors of flexible bodies.
}%
{%
  本研究の実験結果について考察する.
  まず, シミュレーション実験では確率な振る舞いをする簡単な系において, 系のパラメータを明示的に与えなくても, 得られた動作データからparametric biasによってその際のダイナミクスが綺麗に自己組織化可能なことがわかった.
  また, 動作データからオンラインでparametric biasを更新することで, 現在の系の状態を認識することができる.
  $\bm{p}$にはその系の特徴が詰まっており, SPNPBによる推論から, 狙った通りの予測動作遷移, 分散が出力されることがわかった.
  これらを用いて分散最小化を含んだ学習制御を行うことで, 分散が高く不安定な動作を避けながらタスクを実現できることがわかった.
  次に, Fetchにおける実験では, 実機においても, 床の摩擦やテクスチャ等の暗黙的なパラメータをparametric biasによって自己組織化可能なことがわかった.
  オンライン学習により床の材質が変わったことを検知し, 自身の動作する環境を認識することができる.
  最後に, これまでよく動作がスタックしてしまった回転後の後進に本研究を適用し, スタックしないような動作が最適化により生成されることがわかった.
  直接後進すると動作がスタックするのに対して, 一度前進してキャスターの向きを変化させ, 回転方向に動き続けることでスタックフリーな動作が生成された.
  また, これは$\bm{p}$を変化させることで振る舞いが変化し, Corridorにおける$\bm{p}$を使用した場合は, 回転しながら無理やり後進を続ける動作が散見された.
  Corridorは摩擦の少ない床であるため一度前進しなくてもスタックしにくいという特徴が影響したためだと考えられる.
  本実験では適切ではない$\bm{p}$を利用しても大きく性能に差は出なかったが, 他の動作では影響が出る場合も考えられる.

  本研究の問題点は大きく2つあると考える.
  一つ目は, 最適化に時間がかかるため, 現状では実行周期をあまり高く設定できないことにある.
  通常であれば$w^{ref}=w^{ref, orig}$であるため一瞬で応答ができるが, $w^{ref}$を最適化するのにかかった時間だけ応答が遅れてしまうこと(今回だと200 msec).
  最適化された$w^{ref}$を一度のforwardingで出力するような別のネットワークを作ることが考えられる.
  二つ目は, 動作の確率的な振る舞いを正規分布としてモデル化した点である.
  本実験ケースも厳密には正規分布で近似できるようなものではなく, 2種類の動きから確率的に動作が選ばれる等, 別の系では適用できなくなる可能性がある.
  今後様々な系に本研究を適用し, その有効範囲を調べる必要がある.

  本研究では台車ロボットを例に扱ったが, より複雑でモデル化の難しい, 筋骨格ヒューマノイドやSoft Robotへの適用が今後期待される.
  柔軟身体のダイナミクス変化・確率的振る舞いのモデル化に利用していきたい.
}%

\section{Conclusion} \label{sec:conclusion}
\switchlanguage%
{%
  In this study, we proposed a stochastic predictive network with parametric bias, SPNPB, which includes stochastic behavior and implicit environmental information, and developed an adaptive robot control system including variance minimization using SPNPB.
  SPNPB can embed stochastic behaviors to the recurrent neural network by outputting mean and variance, and can embed environmental information by using parametric bias from the motion data in various environments.
  We updated the parametric bias with environmental information online, designed a loss function that executes a task while minimizing the variance of the motion, and calculated the control command by backpropagation technique and gradient descent method.
  By using this study, we have shown that it is possible to minimize the variance of the motion and to execute more stable motion while adapting to the current surrounding environment, both in simulation and in the actual mobile robot Fetch.
  In the future, we would like to apply this method to robots with flexible bodies and more complex structures to confirm its effectiveness.
}%
{%
  本研究では, 確率的挙動・暗黙的な環境情報を含んだ予測モデル型のニューラルネットワーク SPNPBを提案し, これを用いた分散最小化を含む環境適応型ロボット制御を開発した.
  予測モデルのrecurrent neural networkに平均と分散を出力させることで確率的挙動を埋め込み, 様々な環境における動作データからparametric biasを自己組織化させることで環境情報を埋め込むことができる.
  環境情報を含むparametric biasをオンラインで更新し, 出力の分散を最小化しつつtaskを実行する損失関数を設計して誤差逆伝播と勾配降下により制御入力を計算した.
  これにより, 現在の動作環境に適応しつつ, 動作の分散を最小化してより安定的な動作が実行可能であることを台車型ロボットのシミュレーション・実機において示した.
  今後は, 本手法を柔軟身体を持つロボット・より複雑なロボットに適用し, その有効性を確認していきたい.
}%

{
  \bibliographystyle{IEEEtran}
  \bibliography{main}
}

\end{document}